\documentclass{article}
\usepackage{spconf,amsmath,graphicx,hyperref}

\usepackage{newfloat}
\usepackage{listings}
\usepackage{pifont}
\usepackage{bigstrut}
\usepackage{multirow}
\usepackage{url}            
\usepackage{booktabs}       
\usepackage{amsfonts}       
\usepackage{nicefrac}       
\usepackage{microtype}      

\usepackage[numbers]{natbib}
\usepackage{algorithm}
\usepackage{algorithmic}   
\usepackage{graphicx}       
\usepackage{subcaption}   

\usepackage{booktabs}        
\usepackage{siunitx}         
\usepackage{threeparttable}  
\usepackage{adjustbox}  
\usepackage{multirow}
\usepackage{multicol}

\usepackage[table]{xcolor}
\usepackage{adjustbox}

\usepackage{amssymb}    

\title{Enhancing Layer Attention Efficiency through Pruning Redundant Retrievals}
\name{Hanze Li, Yaosong Du, Zhibo Yao, Mengyao Zeng, Xiuqi Ge, Xiande Huang\sthanks{Corresponding author}}
\vspace{-0.3cm}
\address{De Artificial Intelligence Lab
         }

\begin{document}
\ninept
\maketitle
\begin{abstract}
Growing evidence highlights the impact of layer attention mechanisms in deep neural networks, which enhance inter-layer interactions and representational ability. However, existing layer attention methods suffer from redundancy, as adjacent layers often learn highly similar attention weights. This redundancy leads to multiple layers retrieving nearly identical features, diminishing the model's representational capacity and increasing training time. To address this issue, we propose an Efficient Layer Attention (ELA) architecture composing two key components: first, we quantify redundancy by leveraging Kullback-Leibler (KL) divergence between adjacent layers; second, we introduce an Enhanced Beta Quantile Mapping (EBQM) algorithm to accurately detect and skip redundant retrieving, ensuring model stability while boosting efficiency. Experimental results demonstrate that ELA not only enhances overall performance but also reduces training time by 30\%, yielding improvements in tasks such as image classification and object detection.
\end{abstract}

\section{Introduction}
Numerous studies have demonstrated that enhancing inter-layer interactions in deep convolutional neural networks (DCNNs) can substantially improve performance across various tasks. For instance, ResNet \cite{he2016deep} introduced skip connections to facilitate gradient flow, effectively mitigating the issue of performance degradation in very deep networks. Building upon this idea, DenseNet \cite{huang2017densely} further strengthened layer interactions by reusing information from all previous layers. In addition to these innovations, attention mechanisms have also played a crucial role in boosting model performance. Channel attention mechanisms \citep{hu2018squeeze,wang2020eca,qin2021fcanet,ouyang2023efficient}, refine feature selection by prioritizing the most informative channels, enabling the network to focus on the salient feature maps. Complementarily, spatial attention mechanisms \citep{woo2018cbam,wang2018non,li2023scconv,wang2024multi} improve the network's ability to model spatial hierarchies by directing attention to the most relevant spatial regions, thereby refining the learned representations.

Recently, enhancing inter-layer interactions through layer attention mechanisms has emerged as an effective strategy for improving model's representational ability. DIANet \cite{huang2020dianet} introduced an improved LSTM block along the depth of the network to facilitate interactions between layers. Similarly, \citep{zhao2021recurrence,li2024efficient} employ a lightweight recurrent aggregation module to effectively reuse information from previous layers. MUDD \cite{xiao2025muddformer} enhances layer interaction through dynamic connection weights, which depends on each sequence position in a Transformer block. To further enhance cross-layer interaction, MRLA \cite{fang2023cross} introduced layer attention that leverages a multi-head attention across layers to align features from current layer with those from all preceding layers. Building upon this, \cite{wang2024strengthening} argued that the static nature of MRLA limits efficiency and introduced Dynamic Layer Attention (DLA) to restore dynamic context representation for more efficient layer attention.

\begin{figure}[t]
    \centering
    \includegraphics[width=\linewidth]{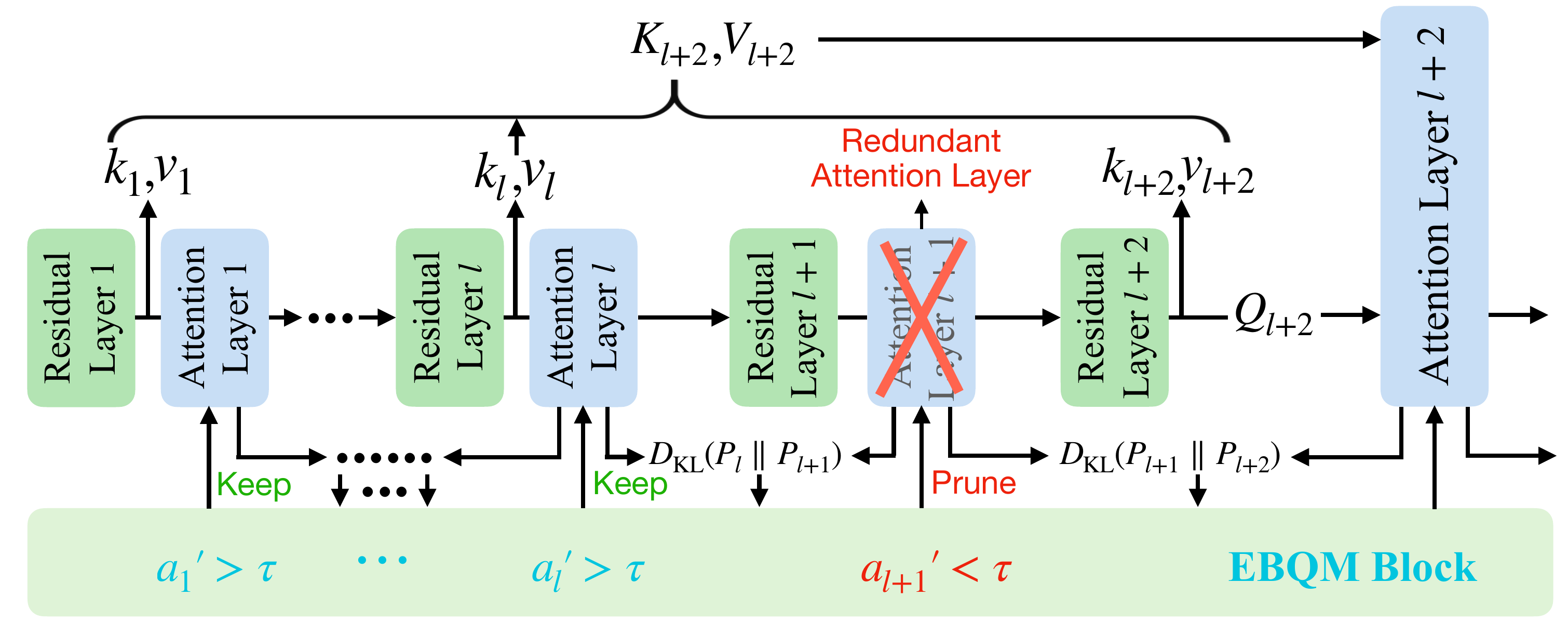}
    \caption{\textbf{Framework of our ELA:} First, the KL divergences of attention scores learned by adjacent layers are computed. These divergences are then fed into the EBQM block and mapped into a target distribution, yielding normalized importance scores $a'_1, \ldots, a'_l, a'_{l+1}$. Finally, attention layers with $a'_l < \tau$ are pruned.
}
    \label{fig:overall}
    \vspace{-0.5cm}
\end{figure}

\begin{figure}[t]
    \centering
    \begin{subfigure}[b]{0.49\linewidth}
        \centering
        \includegraphics[width=\linewidth]{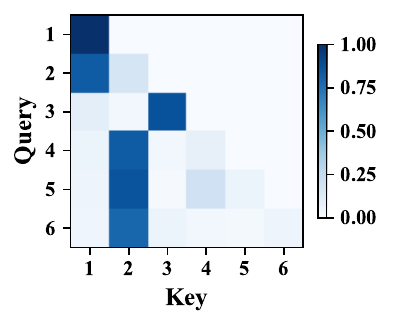}
        \caption{}
        \label{fig:attmap}
    \end{subfigure}
    \hfill
    \begin{subfigure}[b]{0.49\linewidth}
        \centering
        \includegraphics[width=\linewidth]{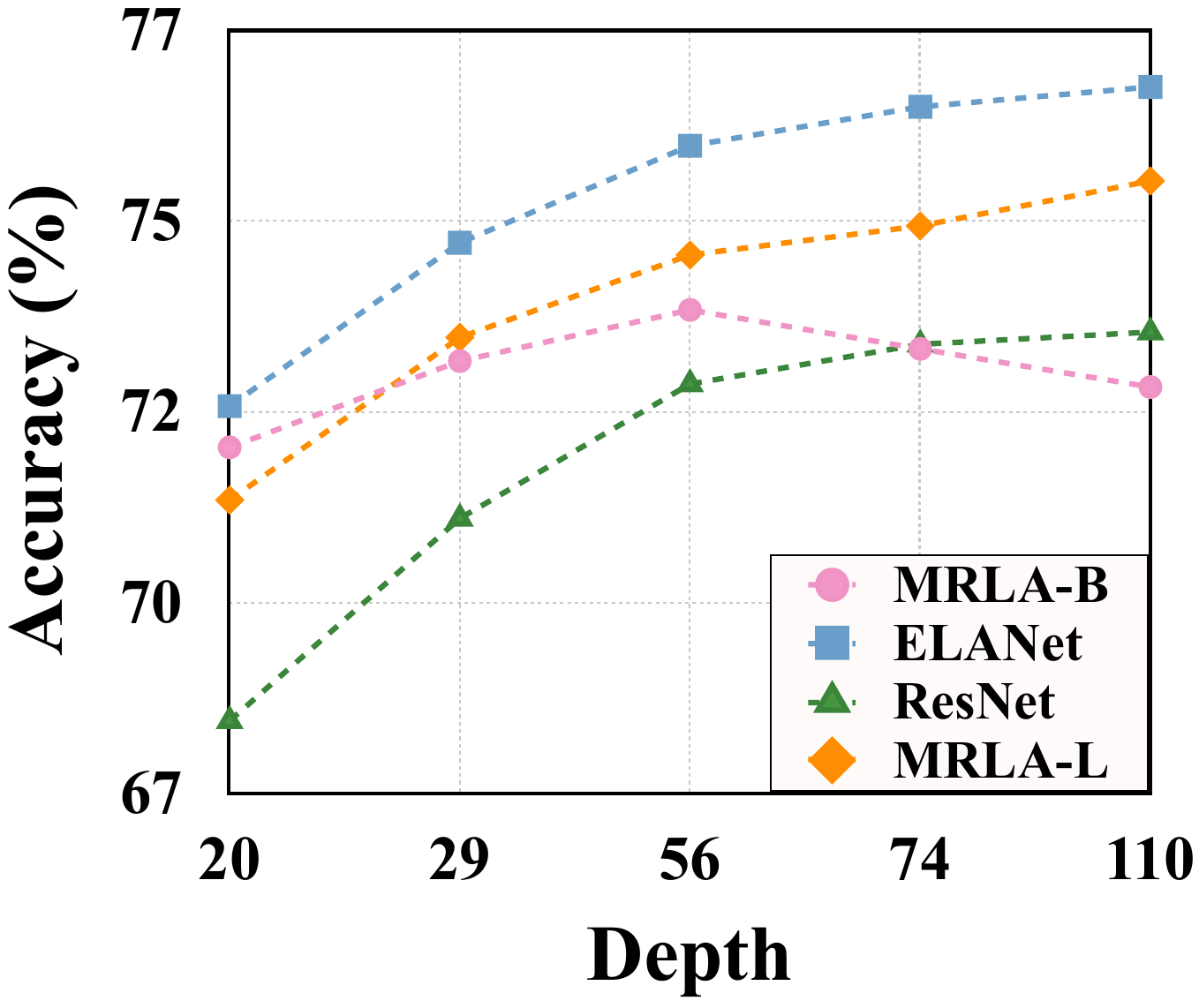}
        \caption{}
        \label{fig:acc_intro2}
        
    \end{subfigure}
    
    \caption{(a) Visualization of attention scores with six adjacent layers in layer attention for ResNet-56 (stage 2). (b) Top1-accuracy comparison of MRLA-B, MRLA-L, ELANet (with ResNet as the backbone) and ResNet at varying depths.}
    \label{fig:cifar_acc}
    \vspace{-0.5cm}
    
\end{figure}

However, we have identified a significant drawback in existing layer attention mechanisms: the redundant retrieval for previous feature across layers. As shown in Figure \ref{fig:attmap}, the attention weights learned by three consecutive layers (the 4th, 5th, and 6th) are nearly identical, all heavily focusing on the 2nd layer. This consistent retrieval pattern is not merely a coincidence—it indicates a systemic redundancy in the attention mechanism. When multiple layers repeatedly attend to the same previous features, the model wastes computation and fails to fully exploit the expressive potential of deeper layers. This issue becomes particularly evident in practice.  As shown in Figure \ref{fig:acc_intro2}, when the network depth exceeds 56 layers, the performance of MRLA-B declines as the depth increases, and with ResNet-110 as the backbone, its accuracy even drops below that of the original ResNet-110 baseline. Although MRLA-L maintains performance gains as the network deepens, the improvement slows down, primarily due to the reduced training efficiency caused by redundant attention patterns. 

Based on these observations, we propose a key research question: \textbf{Is it necessary to relearn similar attention patterns in each layer?} Instead of repeatedly retrieving the same earlier features, later layers—such as the 5th and 6th—can skip the layer attention operation entirely when redundancy is detected, since the relevant information has already been retrieved by the preceding layer. This not only reduces computational overhead, but also alleviates the issue of over-representing certain features, leading to more efficient and balanced information flow across the network.

\begin{figure}[t]
    \centering
    \includegraphics[width=1.0\linewidth]{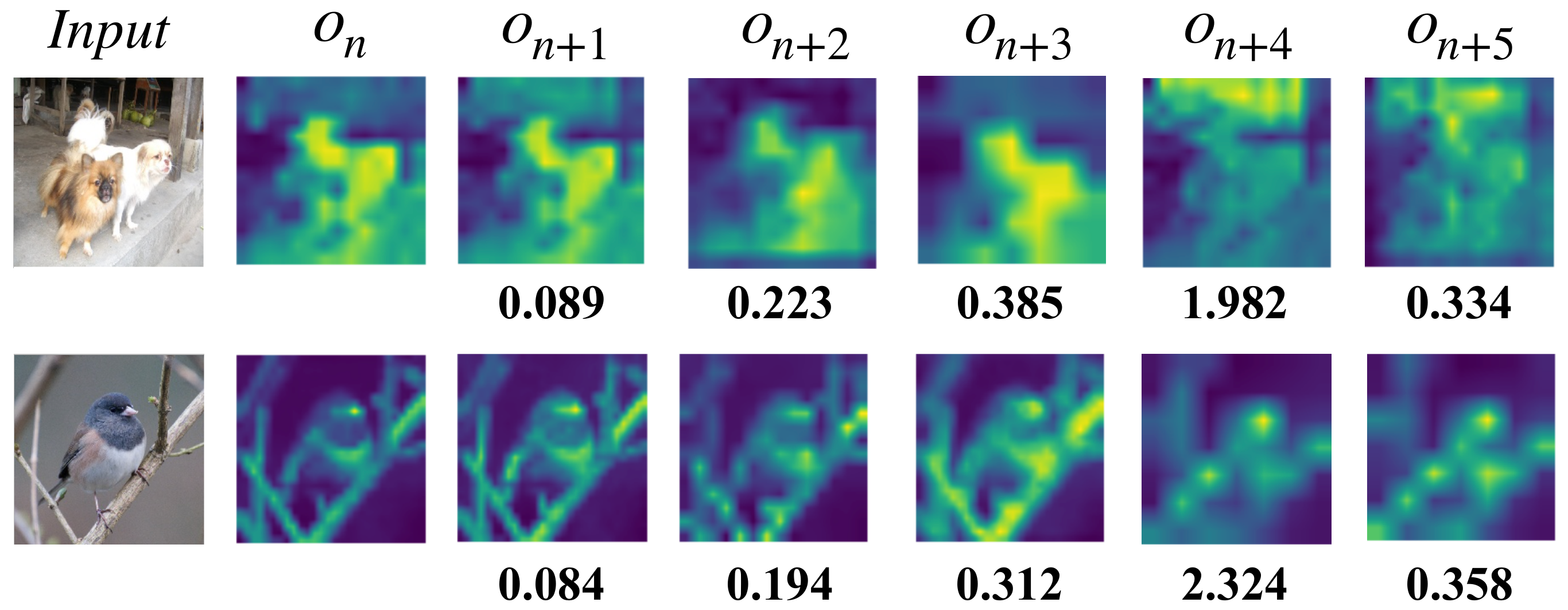}
    \caption{Visualization of attention outputs from six consecutive layers on ResNet-101. The values below each attention output (\(o_t\)) represent the KL divergence of the attention weights between the \(t\)-th and \(t-1\)-th layers. When the KL divergence between adjacent layers' attention distributions is low, their attention outputs also exhibit high visual similarity.} 
    
    \label{fig:visualization}
    \vspace{-0.5cm}
    
\end{figure}

To address this issue, we propose an Efficient Layer Attention (ELA) architecture consisting of two key components: \ding{182} a KL divergence-based strategy to quantify attention redundancy, and \ding{183} an Enhanced Beta Quantile Mapping (EBQM) algorithm to guide the pruning of redundant retrievals. Specifically, we treat attention weights in each layer as probability distributions and compute the KL divergence between successive layers. A low KL divergence indicates that the attention behavior of the current layer is highly similar to that of the previous layer, as shown in Figure \ref{fig:visualization}, suggesting that we can directly discard the retrieval from current layer. However, KL divergence may occasionally fluctuate during training, making it unreliable for pruning in isolation. To address this, EBQM calibrates these divergences into a stable mapping that enables reliable identification of redundant layers. Together, these components allow ELA to reduce unnecessary computation while maintaining or even improving model accuracy.

\vspace{-0.2cm}
\section{Method}
We start by revisiting the layer attention mechanism, followed by the motivation for our proposed Efficient Layer Attention (ELA) architecture. KL divergence is roughly employed to measure the redundant retrievals when adjacent layers attend to similar earlier layers. Then, we propose the Enhanced Beta Quantile Mapping (EBQM) algorithm, which effectively smooths the KL divergence distribution to detect and skip redundant retrievals. 

\subsection{Revisiting Layer Attention}

Layer attention \cite{fang2023cross} enhances inter-layer interactions by employing attention mechanisms, enabling each layer to selectively retrieve relevant information from all preceding layers. Given that $\boldsymbol{x}^l\in\mathbb{R}^{C\times H\times W}$ represents the feature of \( l \)-th layer, where \(C\), \(H\), and \(W\) denote the number of channels, height, and width, respectively, we first extract the query, key, and value representations as follows:
\vspace{-0.2cm}
\begin{align}
\left\{
\begin{array}{ll}
    \boldsymbol{q}^l &= f_q^l(\boldsymbol{x}^l) \\
    \boldsymbol{K}^l &= \mathrm{Concat}\left[ f_k^1(\boldsymbol{x}^1), \dots, f_k^l(\boldsymbol{x}^l) \right] \\
    \boldsymbol{V}^l &= \mathrm{Concat}\left[ f_v^1(\boldsymbol{x}^1), \dots, f_v^l(\boldsymbol{x}^l) \right]
\end{array}
\right.
\vspace{-0.2cm}
\label{eq:1}
\end{align}

Here, \( f_q^l \) extracts the query \(\boldsymbol{q}^l\) from $x^l$, while \( f_k^i \) and \( f_v^i \) extract the key \(\boldsymbol{k}^i\) and the value \(\boldsymbol{v}^i\) from the \(i\)-th layer, respectively. The full key and value matrices \(\boldsymbol{K}^l\) and \(\boldsymbol{V}^l\) are constructed by concatenating the representations from all preceding layers up to \(l\). Then the layer attention can be calculated as:
\begin{align}
    \boldsymbol{o}^l &= \boldsymbol{q}^l (\boldsymbol{K}^{l})^\mathsf{T} \boldsymbol{V}^{l} = \sum_{i=1}^l \boldsymbol{q}^l [f_k^i(\boldsymbol{x}^i)]^\mathsf{T}f_v^i(\boldsymbol{x}^i)
    \label{eq:(2)}
\end{align}
Where \(\boldsymbol{o}^l\) represents the \(l\)-th attention output. For clarity, the softmax and the scaling factor \( \boldsymbol{D}_k \) are omitted here. By substituting:
\begin{align}
    \boldsymbol{q}^l &= \boldsymbol{\lambda}_o^l \odot \boldsymbol{q}^{l-1}
    \label{eq:(3)}
\end{align}
into Equation(\ref{eq:(2)}), a lightweight version, referred to as MRLA-L, can be derived:
\begin{align}
    \boldsymbol{o}^{l} &= \boldsymbol{\lambda}_{o}^{l} \odot \boldsymbol{o}^{l-1} + \boldsymbol{q}^l [f_k^l(\boldsymbol{x}^l)]^\mathsf{T}f_v^l(\boldsymbol{x}^l)
\end{align}
Where \(\boldsymbol\lambda_o^l\) is a learnable vector to dynamically adjust the contribution of prior attention output.


\subsection{Preliminary Knowledge}
\textbf{KL Divergence}
We first utilize the Kullback–Leibler (KL) divergence to roughly approximate the similarity between attention weights learned in adjacent layers of a neural network. Given two discrete probability distributions \(P = \{p_1, \dots, p_n\}\) and \(Q = \{q_1, \dots, q_n\}\), the KL divergence is defined as:
\begin{align}
    D_{\text{KL}}(P \parallel Q) = \sum_{i} p_i \log \left( \frac{p_i}{q_i} \right)
\end{align}
The KL divergence measures the difference between two distributions: a smaller value implies higher similarity, while a larger value indicates greater divergence.

\noindent\textbf{KL Divergence in Layer Attention}
The output of the \(l\)-th layer \(\boldsymbol{x}^l\) can be expressed as a linear combination of value vectors:
\begin{align}
\boldsymbol{x}^l = p_1 v_1 + \dots + p_l v_l
\end{align}
where \(v_i = f_v^i(\boldsymbol{x}^i)\) is the $i$-th value in \(\boldsymbol{V}^{l}\), and \(P = \{p_1, \dots, p_l\}\) denotes the attention weight from \(l\)-th layer, which can be interpreted as a probability distribution.

Similarly, the output of the \(l+1\)-th layer \(\boldsymbol{x}^{l+1}\) can be written as:
\begin{align}
\boldsymbol{x}^{l+1} = p_1' v_1 + \dots + p_{l+1}' v_{l+1}
\end{align}
where \(P' = \{p_1', \dots, p_{l+1}'\}\) is the attention weight derived from \(l+1\)-th layer. Notably, the distribution \(P'\) contains one additional term, $p_{l+1}'$, compared to \(P\).

To enable a valid comparison between the two distributions, we pad \(P\) with an infinitesimal value \(\epsilon\) (where \(\epsilon \to 0\)), resulting in \(P = \{p_1, \dots, p_l, \epsilon\}\), so that its dimensionality matches that of \(P'\).

The KL divergence between the two distributions is then computed as:
\vspace{-0.2cm}
\begin{align}
D_{\text{KL}}(P \parallel P') = \sum_{i=1}^{l} p_i \log \left( \frac{p_i}{p_i'} \right) + \epsilon \log \left( \frac{\epsilon}{p_{l+1}'} \right)\label{eq:8}
\end{align}
In a layer attention network, the previous values \(v_1, \dots, v_{l+1}\) remain consistent between layers \(l\) and \(l+1\). Consequently, the difference between the outputs \(\boldsymbol{x}^l\) and \(\boldsymbol{x}^{l+1}\) is solely attributed to the change in their corresponding attention weight distributions, \(P\) and \(P'\). Therefore, the KL divergence between \(P\) and \(P'\) directly captures the degree of similarity between the two outputs.

\subsection{Efficient Layer Attention}
\noindent\textbf{ELA Architecture}
We present the details of our proposed Efficient Layer Attention (ELA) architecture, as shown in Figure \ref{fig:overall}. Firstly, we compute the KL divergences between attention weights learned from all adjacent layers using Equation (\ref{eq:8}), and collect the results into a divergence vector 
\(\boldsymbol{a}=(a_1,\dots,a_{L-1})\). This vector is then passed through the EBQM module \(\mathcal{E}(·)\), which smooths the divergence values as follows:
\begin{equation}
\boldsymbol{\tilde{a}}
=\mathcal{E}(\boldsymbol{a})
=\bigl(\tilde{a}_1,\dots,\tilde{a}_{L-1}\bigr)
\label{eq:ebqm}
\end{equation}

\noindent
Next, we apply a pruning mechanism based on a pre-defined threshold \(\mu\). A binary mask 
\(\boldsymbol{m} = (m_1, \dots, m_L)\) is generated as:

\vspace{-0.4cm}
\begin{align}
\left\{
\begin{array}{ll}
\boldsymbol{m_{1}}   &= 1 \\[1pt]
\boldsymbol{m_{l+1}} &= 1\!\bigl[\tilde{a}_l \ge \mu\bigr]\in\{0,1\}, 
\quad l = 1,\dots,L-1
\end{array}
\right.
\label{eq:mask-generation}
\end{align}

This mask determines whether the attention outputs from each layer should be preserved or pruned. In each layer \(l\), the representations of query, key, and value are then updated as follows:
\vspace{-0.1cm}
\begin{align}
\left\{
\begin{array}{ll}
    \boldsymbol{q}^{\,l} &= f_q^{\,l}(\boldsymbol{x}^{\,l}) \\[2pt]
    \boldsymbol{K}^{\,l} &= \mathrm{Concat}\!\Bigl[
        m_1\,f_k^{\,1}(\boldsymbol{x}^{\,1}),\;
        \dots,\;
        m_l\,f_k^{\,l}(\boldsymbol{x}^{\,l})
    \Bigr] \\[4pt]
    \boldsymbol{V}^{\,l} &= \mathrm{Concat}\!\Bigl[
        m_1\,f_v^{\,1}(\boldsymbol{x}^{\,1}),\;
        \dots,\;
        m_l\,f_v^{\,l}(\boldsymbol{x}^{\,l})
    \Bigr]
\end{array}
\right.
\vspace{-0.1cm}
\label{eq:1}
\end{align}

Then refined query \(\boldsymbol{q}^{\,l}, \boldsymbol{K}^{\,l}\) and \(\boldsymbol{V}^{\,l}\) are then used to compute the layer attention for layer \(l\) as described in Equation~(\ref{eq:(2)}). Here we set a threshold instead of fixing a pruning ratio, since the number of layers to prune may vary across epochs within the same training run, and enforcing a static ratio would undermine robustness.


\subsection{EBQM Algorithm}
To solve the problem of fluctuations of KL divergences, we propose a novel \textbf{Enhanced Beta Quantile Mapping (EBQM)} algorithm to smooth the distribution of KL divergence. Initially, we retain only the KL divergence values below the $\gamma$-quantile, which corresponds to the lower $\gamma$ proportion of the values (set to $\gamma=0.9$ in our experiments), in order to mitigate the effect of large outliers on normalization, and then scale these values to the range $x \in [0,1]$ to facilitate thresholding. However, these normalized values are often tightly clustered, making the pruning highly sensitive to small variations across training runs. For example, with a threshold \( \mu =0.3 \), six layers with KL divergences between 0.31 and 0.33 in one run may shift to between 0.27 and 0.29 in another. This results in up to six fewer skipped layers in the latter case, introducing inconsistencies that can significantly impact model performance and robustness.

\begin{algorithm}[htbp]
    \caption{Enhanced Beta Quantile Mapping (EBQM)}
    \label{alg:layerskip}
    \textbf{Input}: KL divergences from layer 1 to layer \( l \): \( [a_1, a_2, \dots, a_l] \)\\
    \textbf{Parameter}: Quantile \( \gamma \), Beta distribution parameters \( \alpha, \beta \), number of epochs \( E \)\\
    \textbf{Output}: Updated KL divergences \( [\tilde{a}_1,\dots,\tilde{a}_{L-1}\bigr] \) with skipped layers
    \begin{algorithmic}[1] 
        \FOR {epoch \( e = 1 \) to \( E \)}
            \STATE Sort KL divergences in ascending order: 
            \vspace{-0.1cm}
            \[[a_1,a_2, \dots, a_l] \rightarrow [a_{(1)}, a_{(2)}, \dots, a_{(l)}] \]\\
            \vspace{-0.4cm}
            \[\quad a_{(1)} \leq a_{(2)} \leq \dots \leq a_{(l)}\]
            \vspace{-0.4cm}
            \STATE Extract \( \gamma \)-quantile: 
            \[
            S_\gamma = \{a_i \mid a_i \leq Q_\gamma\},\quad Q_\gamma = \gamma \text{-quantile of } [a_1, \dots, a_l]
            \]
            \vspace{-0.4cm}
            \FOR {each \( a_j \in S_\gamma \)}
                \STATE Normalize \( a_j \):
                \vspace{-0.2cm}
                \[
                a'_j = \frac{a_j - \min(S_\gamma)}{\max(S_\gamma) - \min(S_\gamma)}
                \]
                \vspace{-0.2cm}
                \STATE Apply Beta CDF: \( b_j = F_{\text{Beta}}(a'_j; \alpha, \beta) \)
                
            \ENDFOR
        \ENDFOR
        \STATE \textbf{return} Updated KL divergences \( [\tilde{a}_1,\dots,\tilde{a}_{L-1}\bigr] \)
    \end{algorithmic}
\end{algorithm}

To mitigate this issue, we propose preprocessing the normalized KL divergence distribution before thresholding. Drawing inspiration from the Quantile Mapping algorithm—commonly applied in meteorology and hydrology to map raw data onto a target distribution—we adopt the Beta distribution as the target distribution and perform the transformation through its cumulative distribution function (CDF).





The CDF of the Beta distribution is defined as
\vspace{-0.2cm}
\begin{align}
\text{F}_{\text{Beta}}(x) =
\begin{cases} 
0, & x < 0, \\[3pt]
\displaystyle \int_0^x \frac{t^{\alpha - 1}(1 - t)^{\beta - 1}}
{\int_0^1 u^{\alpha - 1}(1 - u)^{\beta - 1}\, du} \, dt, & 0 \leq x \leq 1, \\[6pt]
1, & x > 1.
\end{cases}
\end{align}

\vspace{-0.2cm}
The CDF of the Beta distribution, unlike that of distributions such as the Gamma distribution, is confined to $[0,1]$, which makes it suitable for mapping the normalized KL divergence. By tuning its shape parameters $\alpha$ and $\beta$, clustered values are dispersed, leading to more stable and robust pruning across training stages.

\section{Experiments}

We validate the efficiency and effectiveness of ELA on two major tasks. For image classification, we conduct experiments on both ResNet and Vision Transformer (ViT) backbones. For object detection, we adopt the widely used Faster R-CNN \cite{ren2015faster} and Mask R-CNN \cite{he2017mask} as detectors. In both tasks, ELA is compared against nearly all existing layer interaction methods, demonstrating its superior performance across diverse architectures and applications.
 
\subsection{Experiment of Image Classification Task} \label{sec:imageclassification}
\begin{figure}[htbp]
    \centering
    \begin{subfigure}[b]{0.48\linewidth}
        \centering
        \includegraphics[width=\linewidth]{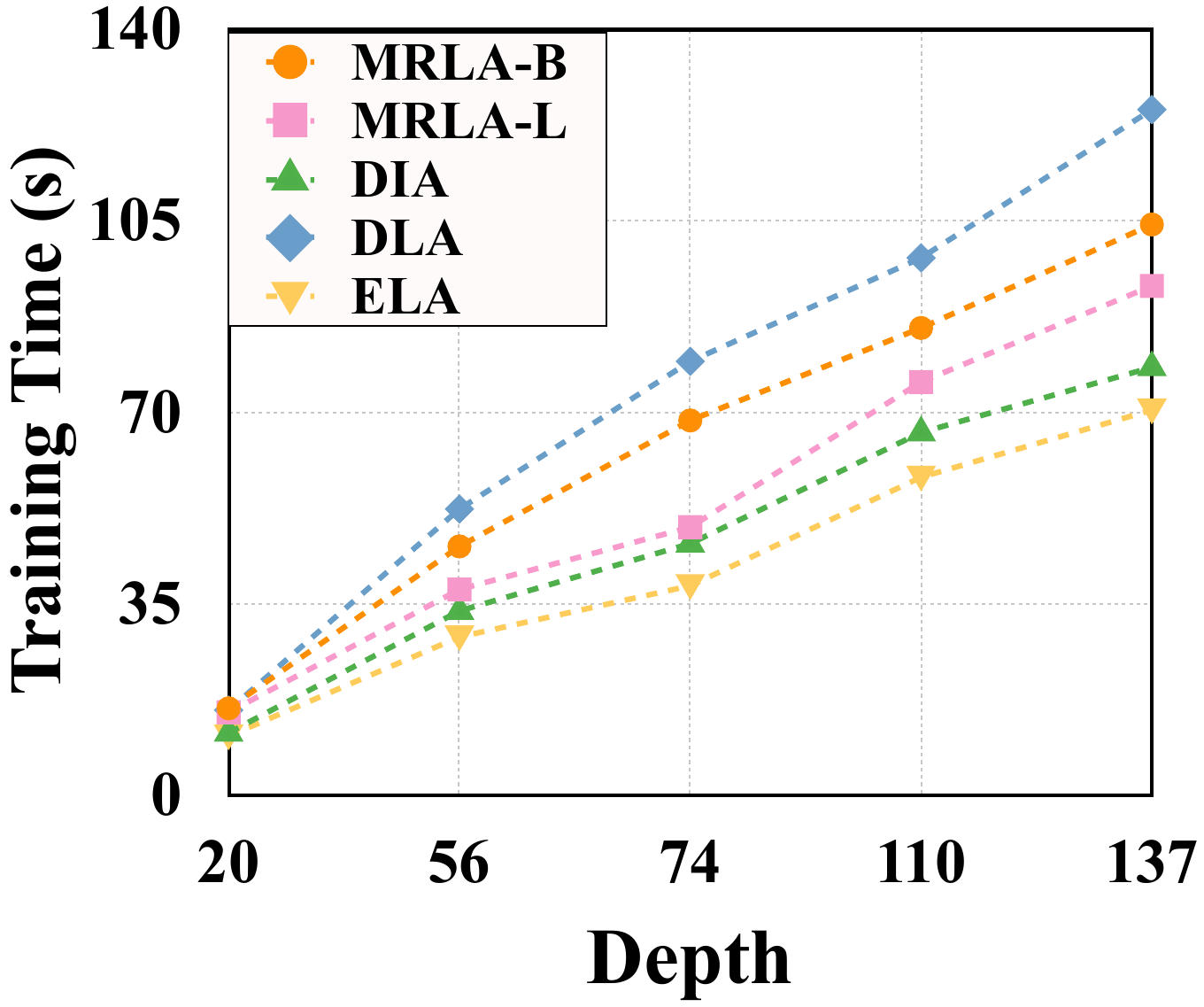}
        \caption{\scriptsize Training Time on CIFAR-10}
        \label{fig:time_10}
    \end{subfigure}
    \hspace{0.01\linewidth}
    \begin{subfigure}[b]{0.48\linewidth}
        \centering
        \includegraphics[width=\linewidth]{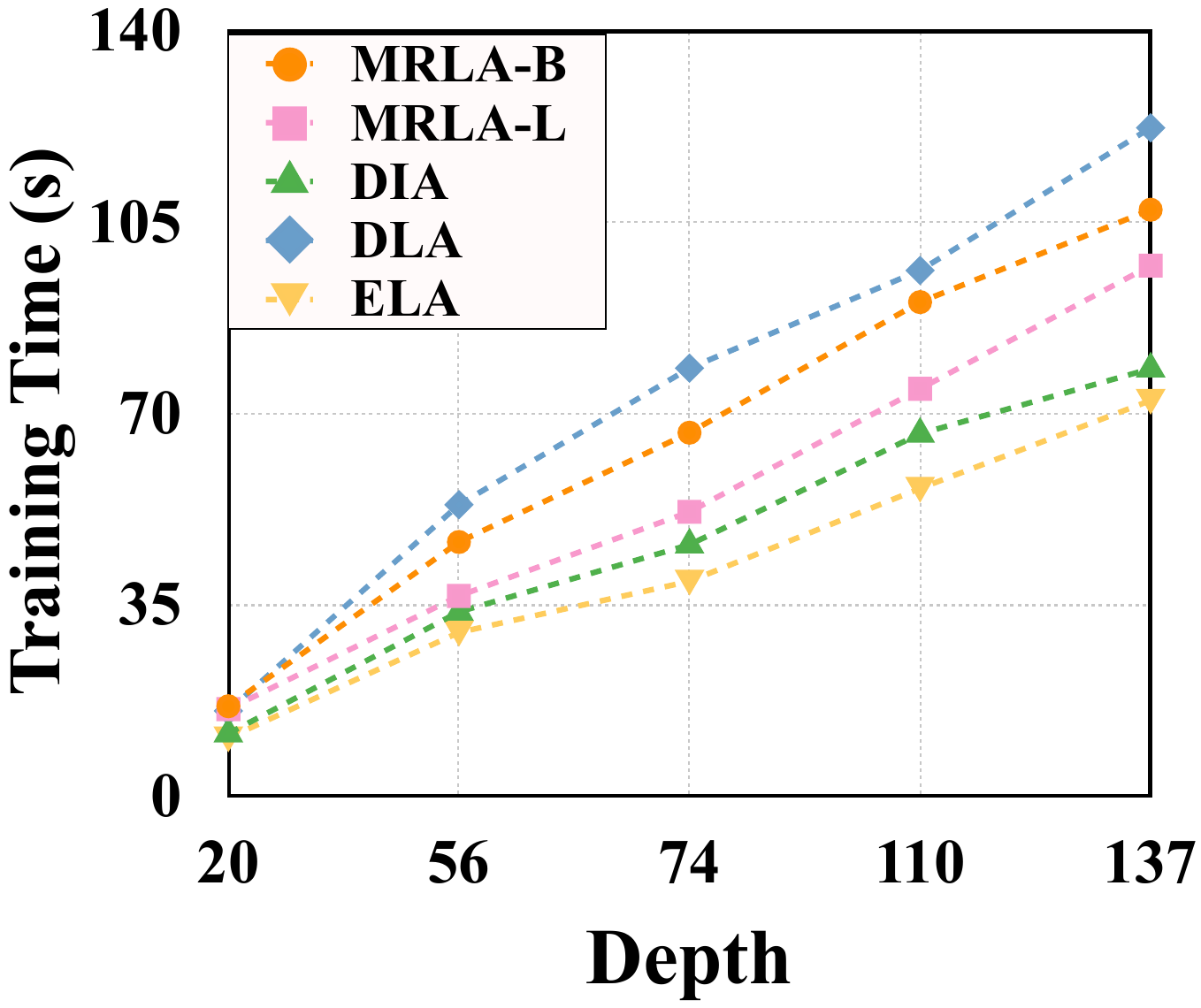}
        \caption{\scriptsize Training Time on CIFAR-100}
        \label{fig:time_100}
    \end{subfigure}
    \caption{Comparison of per-epoch training time (s) for layer interaction models at different depths under identical training conditions.}
    \label{fig:time_cifar}
\end{figure}

We carried out experiments on the CIFAR-10, CIFAR-100 datasets, utilizing ResNet \cite{he2016deep} as the backbone compared with various layer interaction methods \citep{he2016deep,huang2020dianet,fang2023cross,wang2024strengthening} and we applied the data augmentation techniques described in \cite{huang2016deep}. We set \(\alpha\) and \(\beta\) to 5 and 1, respectively. ELANet outperforms all baselines with strong robustness, as shown in Table \ref{tab:cifar_acc} and it substantially reduces training cost as shown in Figure \ref{fig:time_cifar}, cutting time by 30–35\% relative to original MRLA-B.

\begin{table*}[t!]
\centering
\caption{Object detection results on the COCO2017 dataset using ResNet-50 backbone with Faster R-CNN and Mask R-CNN detectors. }
\vspace{-0.1cm}
\tiny
\renewcommand{\arraystretch}{0.35}
\setlength{\tabcolsep}{3.4pt}
\resizebox{\textwidth}{!}{%
\begin{tabular}{lcccccc|cccccc}
\toprule
\multirow{2}{*}{\textbf{Method}} 
& \multicolumn{6}{c|}{\textbf{Faster R-CNN}\cite{ren2015faster}} 
& \multicolumn{6}{c}{\textbf{Mask R-CNN}\cite{he2017mask}} \\
\cmidrule(lr){2-7}\cmidrule(lr){8-13}
& \textbf{AP} & \textbf{AP$_{50}$} & \textbf{AP$_{75}$} & \textbf{AP$_S$} & \textbf{AP$_M$} & \textbf{AP$_L$}
& \textbf{AP} & \textbf{AP$_{50}$} & \textbf{AP$_{75}$} & \textbf{AP$_S$} & \textbf{AP$_M$} & \textbf{AP$_L$} \\
\midrule
ResNet \cite{he2016deep}   & 36.4 & 58.2 & 39.2 & 21.8 & 40.0 & 46.2 & 37.2 & 58.9 & 40.3 & 34.1 & 55.5 & 36.2 \\
SE \cite{hu2018squeeze}    & 37.7 & 60.1 & 40.9 & 22.9 & 41.9 & 48.2 & 38.7 & 60.9 & 42.1 & 35.4 & 57.4 & 37.8 \\
RLA \cite{zhao2021recurrence} & 38.8 & 59.6 & 42.0 & 22.5 & 42.9 & 49.5 & 39.5 & 60.1 & 43.4 & 35.6 & 56.9 & 38.0 \\
MRLA-L \cite{fang2023cross}  & 40.1 & 61.3 & 43.8 & 24.0 & 43.9 & 52.2 & 40.4 & 61.8 & 44.0 & 36.9 & 57.8 & 38.3 \\
DLA \cite{wang2024strengthening} & 40.3 & 61.6 & 43.9 & 24.1 & 44.2 & 52.7 & 40.9 & 62.0 & 44.7 & 37.0 & 58.8 & 39.0 \\
S6LA \cite{liu2025layers}   & 40.3 & \textbf{61.7} & 43.8 & 24.2 & 44.0 & 52.5 & 40.6 & 61.5 & 44.2 & 36.7 & 58.3 & 38.3 \\
\textbf{ELA} & \textbf{40.7} & \textbf{61.7} & \textbf{44.4} & \textbf{24.6} & \textbf{44.2} & \textbf{52.9} 
             & \textbf{41.4} & \textbf{62.4} & \textbf{45.2} & \textbf{37.3} & \textbf{59.2} & \textbf{39.8} \\
\bottomrule
\end{tabular}}
\label{tab:object detetion result}
\end{table*}

\subsection{Experiment of Object Detection Task}
In the object detection task, we evaluate our approach on COCO2017 dataset using Faster R-CNN and Mask R-CNN, with all models implemented in the open-source MMDetection toolkit under default configurations. Our ELA outperforms all channel attention \cite{hu2018squeeze} and layer interaction methods as shown in Table \ref{tab:object detetion result}, and achieves the optimal trade-off between accuracy and parameter efficiency, requiring only 0.2M more parameters than the original ResNet backbone.

\begin{table}[htbp]
    \centering
    \caption{Top-1 accuracy (\%) of different models with ResNet backbone on CIFAR-10 and CIFAR-100. Results are reported as mean $\pm$ standard deviation over five runs.}
    \vspace{-0.1cm}
    \renewcommand{\arraystretch}{0.9}
    \resizebox{0.48\textwidth}{!}{%
    \begin{tabular}{@{}lcccccc@{}}
        \toprule
        \textbf{Data}& \multicolumn{2}{c}{\textbf{CIFAR-10}} & \multicolumn{2}{c}{\textbf{CIFAR-100}} \\
          \cmidrule(lr){2-3} \cmidrule(lr){4-5}
         \textbf{Models}& \textbf{Params} & \textbf{Top-1} & \textbf{Params} & \textbf{Top-1 } \\
        \midrule
        \textbf{ResNet-56} \cite{he2016deep} & 0.59M & $93.01\pm0.21$ & 0.61M & $72.36\pm0.31$ \\
        DIA \cite{huang2020dianet}  & 0.81M & $93.56\pm0.22$ & 0.83M & $74.01\pm0.25$ \\
        MRLA-B \cite{fang2023cross} & 0.62M & $93.72\pm0.37$ & 0.64M & $73.89\pm0.29$ \\
        DLA \cite{wang2024strengthening} & 0.80M & $94.52\pm0.22$ & 0.82M & $74.79\pm0.27$ \\
        S6LA \cite{liu2025layers} & 0.71M & $94.09\pm0.25$ & 0.73M & $74.23\pm0.22$ \\
        \textbf{ELA}  & 0.62M & $\mathbf{94.89\pm0.25}$ & 0.64M & $\mathbf{75.48\pm0.24}$ \\
        \midrule
        \textbf{ResNet-110} \cite{he2016deep} & 1.15M & $92.94\pm0.22$ & 1.17M & $73.04\pm0.33$ \\
        DIA \cite{huang2020dianet}  & 1.37M & $94.13\pm0.27$ & 1.39M & $75.14\pm0.24$ \\
        MRLA-B \cite{fang2023cross}& 1.21M & $94.01\pm0.29$ & 1.23M & $74.72\pm0.37$ \\
        DLA \cite{wang2024strengthening} & 1.39M & $94.36\pm0.35$ & 1.41M & $75.12\pm0.24$ \\
        S6LA \cite{liu2025layers}& 1.33M & $93.76\pm0.11$ & 1.35M & $75.31\pm0.26$ \\
        \textbf{ELA}  & 1.21M & $\mathbf{95.08\pm0.31}$ & 1.23M & $\mathbf{76.25\pm0.22}$ \\
        \bottomrule
    \end{tabular}%
    }
    
    \label{tab:cifar_acc}
\end{table}

\vspace{-0.3cm}
\vspace{-0.5cm}

\subsection{Experiment Results on ViT Architecture}

To assess the effectiveness of our ELA mechanismon on ViT architecture, we compare it against original MRLA \cite{fang2023cross} using DeiT \cite{pmlr-v139-touvron21a} and PVTv2 \cite{wang2022pvt} on ImageNet-1k dataset. As shown in Table \ref{tab:s6la_vs_mrla}, ELA consistently achieves the best Top-1 and Top-5 accuracy across both backbones, surpassing MRLA and the baseline. The gains are particularly pronounced on DeiT, where ELA improves Top-1 accuracy from 79.9\% to 81.0\% . These results demonstrate that ELA provides superior generalization and establishes  state-of-the-art performance on transformer architectures.

\begin{table}[htbp]
\centering
\caption{Top-1 and Top-5 accuracy (\%) of DeiT and PVTv2 backbones with MRLA and ELA on ImageNet-1K using ResNet-50.}
\setlength{\tabcolsep}{3pt}
\renewcommand{\arraystretch}{0.9}
\resizebox{0.47\textwidth}{!}{%
\begin{tabular}{llcc|llcc}
\toprule
\multicolumn{4}{c|}{\textbf{DeiT}\cite{pmlr-v139-touvron21a}} & \multicolumn{3}{c}{\textbf{PVTv2}\cite{wang2022pvt}} \\
\cmidrule(lr){1-4}\cmidrule(lr){5-8}
\textbf{Backbone} & \textbf{Method} & \textbf{Top-1} & \textbf{Top-5} 
& \textbf{Backbone} & \textbf{Method} & \textbf{Top-1} & \textbf{Top-5} \\
\midrule
 & ResNet-50 & 72.6 & 91.1 & & ResNet-50 & 70.0 & 89.7 \\
 DeiT-Ti       & +MRLA-L \cite{fang2023cross}    & 73.0 & 91.7 &     PVTv2-B0      & +MRLA-L \cite{fang2023cross}    & 70.6 & 90.0 \\
        & +ELA     & \textbf{73.4} & \textbf{92.1} & & +ELA & \textbf{70.8} & \textbf{90.3} \\
\midrule
  & ResNet-50 & 79.9 & 95.0 &  & ResNet-50 & 78.3 & 94.3 \\
 DeiT-S       & +MRLA-L \cite{fang2023cross}    & 80.7 & 95.3 &    PVTv2-B1      & +MRLA-L \cite{fang2023cross}   & \textbf{78.9} & \textbf{94.9} \\
        & +ELA     & \textbf{81.0} & \textbf{95.8} & & +ELA & \textbf{78.9} & 94.7 \\
\bottomrule
\end{tabular}}
\label{tab:s6la_vs_mrla}
\end{table}

\vspace{-0.5cm}

\section{Ablation Studies}

To evaluate the impact of EBQM hyperparameters, we conduct an ablation on $\alpha$ and $\beta$ with ResNet-50 on CIFAR-100. Results in Table \ref{tab:ebqm_ablation} show consistent gains over the without-EBQM setting, with the best performance achieved at $\alpha=5, \beta=1$. It can be observed that EBQM enhances both the capability and stability of the model. Although $\alpha$ and $\beta$ are tuned manually, the performance remains consistently strong across different settings.

\begin{table}[htbp]
  \centering
  \caption{Top-1 accuracy (\%) for varying \(\alpha\) (with \(\beta=1\)) and varying \(\beta\) (with \(\alpha=5\)) on CIFAR-100 with ResNet-50 backbone. Results are reported as mean $\pm$ standard deviation over five runs.}
  \renewcommand{\arraystretch}{1.0}
  \resizebox{0.48\textwidth}{!}{%
  \setlength{\tabcolsep}{2pt}
  \begin{tabular}{c|cccccc}
    \toprule
    \textbf{Setting} & w/o EBQM & \(\alpha\)/\(\beta\)=1 & \(\alpha\)/\(\beta\)=2 & \(\alpha\)/\(\beta\)=3 & \(\alpha\)/\(\beta\)=4 & \(\alpha\)/\(\beta\)=5 \\
    \midrule
     \(\beta\)=1 &  73.92$\pm$0.41& 74.88$\pm$0.29 & 74.65$\pm$0.33 & 75.01$\pm$0.45  & 75.12$\pm$0.27 & \textbf{75.48$\pm$0.24} \\
     \(\alpha\)=5 & 73.92$\pm$0.41 & \textbf{75.48$\pm$0.24} & 74.44$\pm$0.38 & 75.22$\pm$0.28 & 75.01$\pm$0.31 & 74.73$\pm$0.35  \\
    \bottomrule
  \end{tabular}}
  \label{tab:ebqm_ablation}
\end{table}

\vspace{-0.5cm}
\section{Conclusion}
In this paper, we uncover the inherent redundancy in existing layer attention mechanisms, characterized by the high similarity of attention weights learned by adjacent layers. We further analyze the two key issues caused by this redundancy: degraded model performance and extended training time. To address these challenges, we propose a novel approach that first evaluates the importance of each attention layer using KL divergence and then employs an Enhanced Beta Quantile Mapping Algorithm to prune redundant layers. Experimental results on image classification and object detection tasks demonstrate that our method effectively eliminates redundant information, resulting in a network that outperforms the original layer attention model.

\bibliographystyle{IEEEbib}
\bibliography{icassp}

\end{document}